\documentclass[10pt,twocolumn,letterpaper]{article}

\usepackage{iccv}
\usepackage{times}
\usepackage{epsfig}
\usepackage{graphicx}
\usepackage{amsmath}
\usepackage{amssymb}

\usepackage{bm}
\usepackage{booktabs}
\usepackage{multirow}
\usepackage{xcolor}
\usepackage[accsupp]{axessibility} 
\usepackage[breaklinks=true,bookmarks=false]{hyperref}

\iccvfinalcopy 

\def\iccvPaperID{6661} 

\ificcvfinal\fi

\begin{document}

\title{Identity-Consistent Aggregation for Video Object Detection}

\author{Chaorui Deng$^1$, ~Da Chen$^2$, ~Qi Wu$^{1,*}$\\
$^1$Australia Institute of Machine Learning, University of Adelaide\\
$^2$Department of Computer Science, University of Bath\\
}

\maketitle
\ificcvfinal\thispagestyle{empty}\fi

\makeatletter
\def\blfootnote{\xdef\@thefnmark{}\@footnotetext}
\makeatother

\begin{abstract}
In Video Object Detection (VID), a common practice is to leverage the rich temporal contexts from the video to enhance the object representations in each frame.
Existing methods treat the temporal contexts obtained from different objects indiscriminately and ignore their different identities.
While intuitively, aggregating local views of the same object in different frames may facilitate a better understanding of the object.
Thus, in this paper, we aim to enable the model to focus on the identity-consistent temporal contexts of each object to obtain more comprehensive object representations and handle the rapid object appearance variations such as occlusion, motion blur, \emph{etc}.
However, realizing this goal on top of existing VID models faces low-efficiency problems due to their redundant region proposals and nonparallel frame-wise prediction manner.
To aid this, we propose \textit{ClipVID}, a VID model equipped with Identity-Consistent Aggregation (ICA) layers specifically designed for mining fine-grained and identity-consistent temporal contexts.
It effectively reduces the redundancies through the set prediction strategy, making the ICA layers very efficient and further allowing us to design an architecture that makes parallel clip-wise predictions for the whole video clip.
Extensive experimental results demonstrate the superiority of our method: a state-of-the-art (SOTA) performance (84.7\% mAP) on the ImageNet VID dataset while running at a speed about 7$\times$ faster (39.3 fps) than previous SOTAs.
\blfootnote{$^*$Corresponds to~\texttt{qi.wu01@adelaide.edu.au}. Code is available at \texttt{https://github.com/bladewaltz1/clipvid}.}
\end{abstract}

\section{Introduction}

Video Object Detection (VID) aims to recognize and localize the objects in all frames given a video clip. 
It is a challenging task as it must handle the complex appearance variations of video objects, caused by motion blur, occlusion, rotation, unusual poses, and deformable shapes, \emph{etc}.
To tackle these issues, prior works \cite{feichtenhofer2017detect,kang2017object,kang2016object,zhu2017deep} utilize a set of support frames (\eg, neighboring frames of the target frame), which provide rich temporal contexts, to guide the object detection in the target frame. 
For example, \cite{zhu2017flow,wang2018fully,bertasius2018object} build grid-level relations between the feature maps of the support frames and the target frame to propagate temporal contexts.
More recent SOTA methods \cite{wu2019sequence,chen2020memory,han2020mining,cui2021tf,lin2020dual} adopt object-level relation modules \cite{hu2018relation} to leverage the region proposals extracted from long-range support frames to enhance the object representations in the target frame.

\begin{figure}[t]
  \centering
  \includegraphics[width=1.0\linewidth]{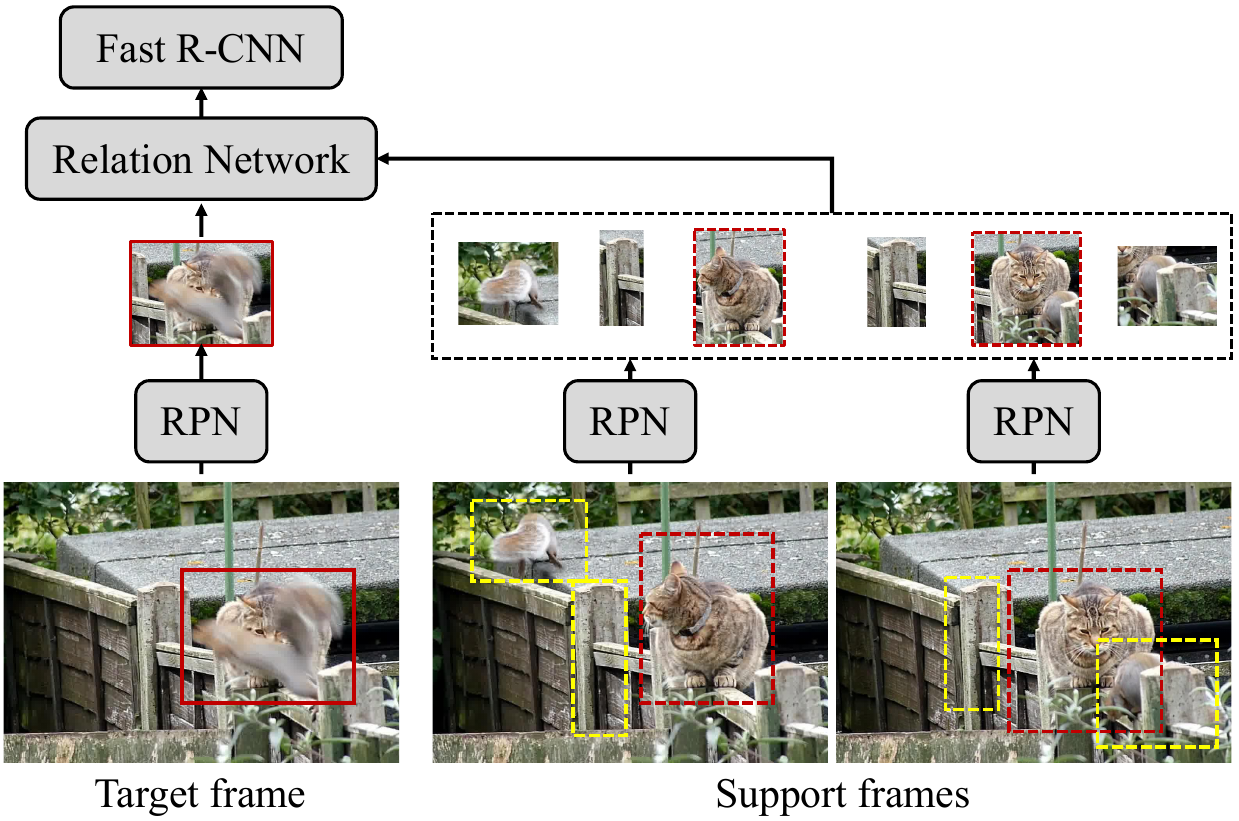}
  \caption{Illustration of the temporal context aggregation in a typical VID method \cite{wu2019sequence}.
  The region proposals from the support frames (dashed boxes) are treated indiscriminately regardless of their object identities.
  However, to detect the cat in the target frame, the region proposals with red dashed boxes should provide more relevant information, as they are obtained from the same cat.
  }  \label{fig:intro}
\end{figure}

However, as shown in Figure \ref{fig:intro}, the temporal contexts from the support frames usually contain irrelevant and noisy information that may negatively affect the object representations.
\emph{E.g.}, when detecting the cat in the target frame, the yellow boxes in the support frames only provide information from different objects and even from the background.
On the other hand, the red boxes are different local views of the same cat, showing it from various perspectives.
Intuitively, incorporating these local views into a unified representation could lead to a more comprehensive understanding of the object, and further facilitate the model to deal with the rapid variations of the object appearance.
Unfortunately, existing methods make no distinction between these two kinds of temporal contexts.
In light of this, we propose an Identity-Consistent temporal context Aggregation (ICA) approach, which aims to discover and utilize the local views of each object to learn its global view to guide the detection.

To achieve this, a prerequisite is to ensure that the region proposals extracted from support frames have a high recall rate for the video objects, 
so that each object in the target frame can find its identity-consistent temporal contexts from them.
This requires existing methods to extract a large number of region proposals (\emph{e.g.}, 300) from each support frame due to the redundant predictions made by their base detectors \cite{dai2016r,ren2016faster}.
However, the computation complexity of the object relation module is usually quadratic to the total number of region proposals in the video.
Therefore, existing methods have to lower the number of region proposals extracted from the support frames to make the computation feasible, decreasing the recall rate of the video objects and hampering the ICA process.
Worse still, the low recall rate also leads to a low detection performance on the support frames.
Thus, given an input video clip, existing methods only make predictions for one target frame, while the support frames are merely used as guidance.
This non-parallel behavior further hampers the model efficiency.

For these reasons, we build our VID model based on the DETR \cite{carion2020end} framework. 
Specifically, instead of generating a large number of redundant object candidates, we represent each object in a video frame with a learnable embedding, \textit{a.k.a.} ``object query'', and iteratively incorporate its object-related visual content from the frame representation and its identity-consistent temporal contexts from the video into the object query.
Hungarian algorithm is used to perform one-to-one bipartite matching between the object queries and the ground-truth objects. 
In this way, the total number of object queries can be typically less than $100$ per frame, which is an order of magnitude smaller than the number of region proposals in previous methods.
Thus, the whole detection process can be achieved efficiently, further allowing our model to perform parallel clip-wise predictions.

The proposed model, termed ClipVID, adopts a clean backbone + Transformer decoder architecture.
It first extracts features from each frame separately using a CNN-based~\cite{he2016deep} backbone.
Then, the object queries for all input frames are adaptively generated and are fed into a Transformer decoder jointly to propagate the temporal contexts.
To perform identity-consistent temporal context aggregation, we assign each object query with an object identity, and additionally predict an identity embedding for it, which is then adopted to select the object queries from other frames that are close in the embedding space. 
These selected object queries are considered to have the same object identity and are fed into an ICA module to maintain the identity consistency of the video objects.
Finally, the object queries from all frames are fed into the detection head jointly to obtain their predictions in parallel.

When evaluated on the ImageNet VID dataset \cite{russakovsky2015imagenet}, ClipVID achieves a significant performance improvement in fast-moving objects, which are the type of objects that suffers mostly from the appearance variations in a video, \emph{e.g.}, motion blur, occlusion, and deformation.
This further leads to a state-of-the-art overall performance (84.7\% mAP) without the need for post-processing.
Moreover, our model is able to run at 39.3 fps, which is about 7$\times$ faster than recent SOTAs.
In summary, our contributions are three-fold:
\begin{enumerate}
    \item We propose the ClipVID model which is able to leverage the identity-consistent temporal contexts to obtain comprehensive representations for the video objects, leading to a SOTA performance on the VID task.
    \item The proposed ClipVID makes clip-wise predictions for the VID task, \emph{i.e.}, detects the objects on all input frames simultaneously, which is significantly faster than previous frame-wise prediction methods.
    \item We conduct extensive experiments to analyze the performance of the proposed ClipVID model.
\end{enumerate}

\section{Related Works}\label{sec:rel}
\noindent\textbf{Images Object Detection}.
Object Detection methods in the image domain have developed rapidly over the years \cite{ren2016faster,girshick2015fast,redmon2016you,dai2016r,liu2016ssd,lin2017feature,lin2017focal,tian2019fcos}.
Among them, Faster RCNN \cite{ren2016faster} is one of the most popular object detectors.
In Faster RCNN, a backbone CNN extracts the image representation and then feeds it into a region-proposing stage to generate a large number of region proposals, 
followed by a detection stage to classify and refine the proposals.
Non-maximum suppression (NMS) is required in both stages to remove redundancy.
Deformable convolution \cite{dai2017deformable} and Relation Network \cite{hu2018relation} are two useful approaches to boost object detection performance.
Specifically, deformable convolution samples feature from dynamic locations to facilitate a more aligned receptive field.
Relation Networks applies self-attention \cite{vaswani2017attention} among the region proposals to enhance their features with contextual information.

\noindent\textbf{Video Object Detection}.
Leveraging temporal contexts as guidance has been proven to be beneficial for frame-wise VID.
In FGFA \cite{zhu2017flow}, estimated flow fields \cite{dosovitskiy2015flownet} are used to wrap the features of neighbor frames to enhance the feature of the target frame.
In STSN \cite{bertasius2018object}, deformable convolutions~\cite{dai2017deformable} are used to sample features from neighbor frames to boost the target frame representation.
STMM \cite{xiao2018video} and PSLA \cite{guo2019progressive} propagate temporal contexts from neighbor frames through a recurrent feature map memory, which communicates with the target frame through grid-level attention.
To acquire more semantically diverse temporal contexts, \cite{wu2019sequence,shvets2019leveraging} adopt long-range support frames as guidance, by feeding the region proposals in the target frame and support frames together into a Relation Network~\cite{hu2018relation}.
MEGA \cite{chen2020memory} and OGEMN \cite{deng2019object} propose memory mechanisms to utilize temporal contexts from both long-range and neighbor support frames. 
HVR-Net \cite{han2020mining} further considers cross-video proposal relations to exploit more diverse temporal cues.
TF-Blender \cite{cui2021tf} inserts a grid-level fusion layer into MEGA to leverage fine-grained and diverse temporal contexts.
In contrast to previous methods where they treat the temporal contexts from different objects indiscriminately, we propose to capture the identity-consistent temporal contexts for each object explicitly.

There have also been attempts to accelerate the frame-wise VID methods through sparse computation \cite{chen2018optimizing,zhu2017deep,zhu2018towards,jiang2020learning}, \emph{i.e.}, assigning more computation budgets (\emph{e.g.}, using heavier backbones or larger frame resolutions) to the keyframes in the video, and assigning fewer computation budgets to the non-key frames.
These strategies can also be applied to our model to further accelerate its speed.
\newline

\begin{figure*}[t]
\begin{center}
\includegraphics[width=0.82\linewidth,height=0.27\textheight]{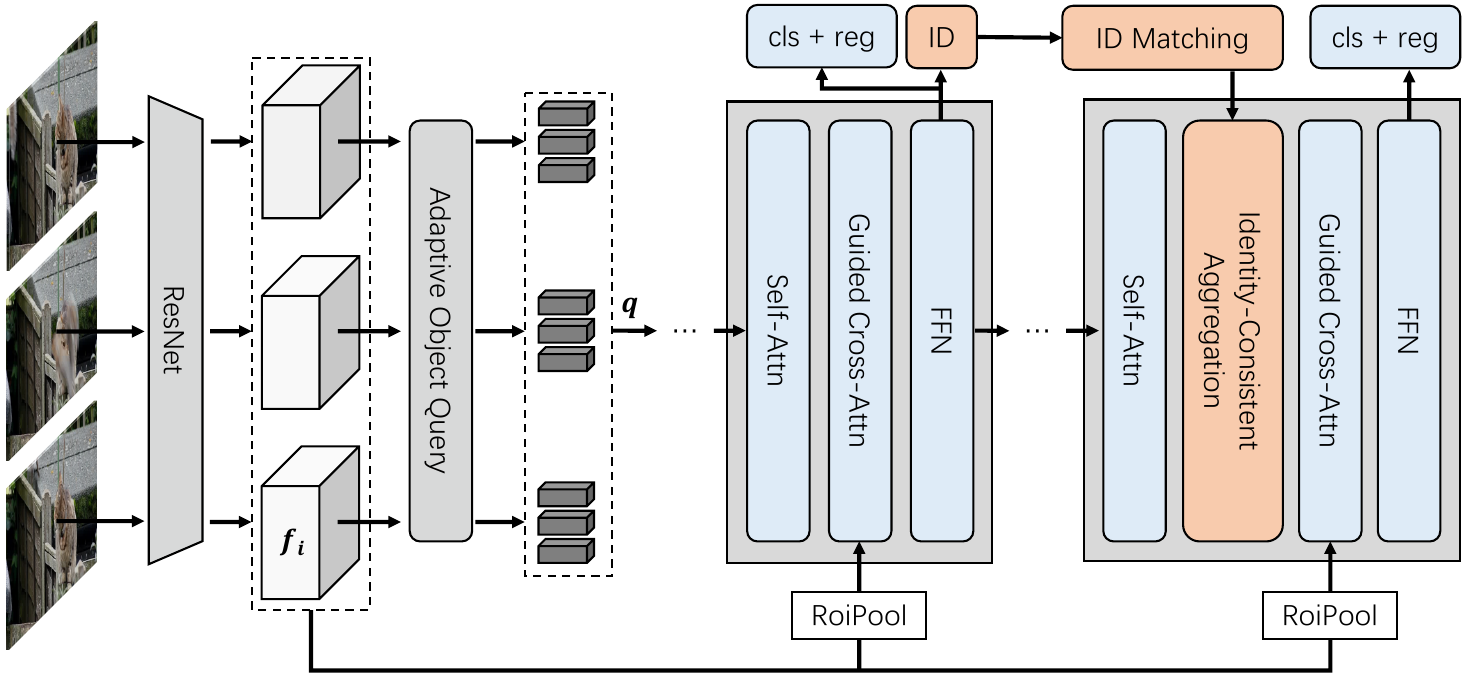}\label{fig:arch}
\caption{The overview of ClipVID. $\bm{f}_i$ and $\bm{q}$ indicate the frame feature and the object queries, respectively.
``cls'', ``reg'', and ``ID'' indicate the classification branch, localization branch, and identity embedding branch of the prediction head, respectively.
Two transformer layers are shown. In the first one, the identity embedding branch predicts the identity embedding of each object query.
Then in the second one, a Matching module is used to group the object queries according to the distances of their identity embeddings, and an ICA module is used to perform identity-consistent temporal context aggregation.
For simplicity, only three frames and three queries per frame are shown.
}
\end{center}
\end{figure*}

\noindent\textbf{End-to-End Detectors for Images/Videos}. The end-to-end object detector DETR \cite{carion2020end} has drawn great attention recently.
The core idea is to use the Hungarian algorithm to perform a one-to-one label assignment between the ground-truth objects and a set of learnable object queries during training.
The DETR framework effectively removes the need for many hand-crafted components like NMS and anchor generation.
The main limitation of DETR is its low convergence speed.
To aid this, \cite{zhu2020deformable,sun2021sparse} adopt guided attention which only selects a subset of locations from the feature map to perform cross attention according to learned reference points/boxes.
Moreover, these reference points/boxes are updated iteratively at each decoder layer. 

The DETR architecture has been applied to the video domain.
TransVOD \cite{he2021end} is a recently proposed DETR-based model which is able to perform end-to-end video object detection. However, it is still a frame-wise prediction method and contains complex sub-modules to process the support and the target frames, respectively.
Differently, the proposed ClipVID is a clean backbone + Transformer decoder architecture that performs clip-wise prediction, which is simple and efficient.
VisTR \cite{wang2021end} for Video Instance Segmentation (VIS) proposes an Instance Sequence Matching strategy that addresses the object linking problem in VIS.
it imposes an assumption that the differences between input frames are mild, which generally does not hold in practice.
Different from VisTR, the proposed ICA can be used to link objects across frames, and is especially effective in dealing with large appearance variations.
\newline

\noindent\textbf{Multi-Object Tracking} (MOT) is also an extensively studied problem in computer vision.
SOTA methods in MOT are dominated by the tracking-by-detection paradigm \cite{chu2019famnet,xu2019spatial,braso2020learning}.
\emph{I.e.}, the objects on each frame are first detected using object detectors like Faster RCNN, then associated together.
In other words, the bounding boxes of these objects are pre-given and the tracker only needs to solve the association task.
Differently, VID focuses on generating high-quality detection results on each frame with the help of temporal contexts.
Some MOT methods \cite{bergmann2019tracking,zhou2020tracking,zhang2021fairmot} perform detection and tracking jointly, where the detected object in previous frames are used to detect and associate the objects in the current frame.
Following this, a more recent work \cite{sun2020transtrack} adopts the DETR framework and uses object queries to detect objects at the current frame and associate them with existing tracklets.
However, the dependence on previous frames enforces these methods to perform frame-wise prediction, which is inefficient. 

\section{Method}
The architecture of the proposed method is shown in Figure~\ref{fig:arch}.
Given the input video clip, the proposed ClipVID first extracts the frame features using a backbone network.
Then, ClipVID generates object queries adaptively for each frame and inputs them into a Transformer decoder where the self-attention operations process object queries from all frames in a unified manner to propagate temporal contexts.
Then in the cross-attention operations, the object queries attend to their corresponding frame features to retrieve object-related visual contents.
In the last several decoder layers, Identity-consistent Aggregation is performed to aggregate identity-consistent temporal contexts into each object query.
A feed-forward network (FFN) with a detection head is applied to all object queries to make parallel clip-wise predictions.

\subsection{Clip-wise Video Object Detection}
ClipVID is an end-to-end video object detector with a backbone + Transformer decoder architecture.
We detail its basic components as follows.
\newline

\noindent\textbf{Backbone}.
Given a video clip of $T$ input frames $\bm{x} = \{\bm{x}_i\}_{i=1}^T$, $ \bm{x}_i \in \mathbb{R}^{H_0 \times W_0 \times 3}$, the backbone model extracts a lower-resolution feature map from each frame, following by a $1\times 1$ convolution layer to reduce its dimension to $d$.
We denote the frame features as $\bm{f} = \{\bm{f}_i\}_{i=1}^T$, where $\bm{f}_i \in \mathbb{R}^{H \times W \times d}$ is for each frame.
Unlike DETR, we do not adopt a transformer encoder to further encode $\bm{f}$ since it is memory-consuming to process the long feature sequences ($T\times H \times W$) of the video clip. 
\newline

\noindent\textbf{Adaptive Object Queries}.
To handle input video clips with arbitrary frame lengths, a naive approach is to share a fixed set of object queries across all frames, which may lead to a low semantic diversity.
Instead, we generate $L$ object queries adaptively for each frame conditioned on a learnable embedding matrix $\bm{e} \in \mathbb{R}^{L\times d}$, resulting in $T \times L$ object queries in total, denoted by $\bm{q} = \{\bm{q}_{ij} \in \mathbb{R}^{1\times d}|i\in [1,T], j\in[1,L]\}$.
Specifically, for $i$-th frame, its $j$-th adaptive object queries $\bm{q}_{ij}$ is obtained by
\begin{equation}
\bm{q}_{ij} = \text{softmax}(\bm{e}_j\bm{m}_i^\top)\bm{m}_i,
\end{equation}
where $\bm{m}_i \in \mathbb{R}^{s^2\times d}$ is obtained by firstly down-sampling the spatial dimensions of $\bm{f}_i$ to $s\times s$ and then flattening its spatial dimensions, and $\bm{e}_j \in \mathbb{R}^{1\times d}$ is the $j$-th embedding.
\newline

\noindent\textbf{Self Attention}. 
The self-attention operation is performed over all $T\times L$ object queries, propagating temporal contexts among all frames to make parallel predictions for the input video clip.
In self-attention, the object query is updated by:
\begin{equation}
\begin{split}
\bm{q}_{ij} \longleftarrow & \text{LN}(\bm{q}_{ij} + \text{MHA}(\bm{q}_{ij}, \bm{q}, \bm{q})), \\
& \forall i \in [1, T], \forall j \in [1, L]
\end{split}
\end{equation}
where $\text{MHA}(q, k, v)$ indicates the multi-head attention~\cite{vaswani2017attention} operation using the query $q$, key $k$, and value $v$.
LN indicates the layer normalization operation \cite{ba2016layer}.
After the self-attention layer, each object query is enhanced with the contextual information from the whole clip.
\newline

\noindent\textbf{Guided Cross Attention}. 
In the cross attention operation, different from DETR where an object query attends to all locations in the feature map, here we adopt the idea of guided attention \cite{zhu2020deformable,sun2021sparse} to accelerate the convergence speed,
\emph{i.e.}, the object query $\bm{q}_{ij}$ only performs cross attention with the locations inside a reference box ${b}_{ij}$.
Specifically, given the frame feature $\bm{f}_i$, RoIPooling \cite{he2017mask} is performed over $\bm{f}_i$ based on ${b}_{ij}$, to extract a $s\times s$ feature map.
The feature map is further flattened along the spatial dimensions, denoted by $\bm{k}_{ij} \in \mathbb{R}^{s^2 \times d}$.
Moreover, we propose to enhance $\bm{k}_{ij}$ with the semantic information from its matched object query through an element-wise adaptation operation:
\begin{equation}\label{eqn:roifeat}
\bm{k}_{ij} \longleftarrow \bm{k}_{ij} + \text{reshape}(\bm{q}_{ij}\bm{W}^p),
\end{equation}
where $\bm{W}^p \in \mathbb{R}^{d\times s^2d}$ is a learnable parameter and $\bm{q}_{ij}\bm{W}^p$ is reshaped to $s^2 \times d$ which is added to $\bm{k}_{ij}$ element-wisely.
This process is important as it fuses the grid-level and instance-level features of an object (obtained from the backbone and the decoder, respectively) in a fine-grained manner, leading to an improved object representation.
Then, cross-attention is performed using $\bm{k}_{ij}$ as the source and $\bm{q}_{ij}$ as the query:
\begin{equation}
\bm{q}_{ij} \longleftarrow \text{LN}(\bm{q}_{ij} + \text{MHA}(\bm{q}_{ij}, \bm{k}_{ij}, \bm{k}_{ij})).
\end{equation}
Through this guided cross-attention, each object query attends to a specific region in the corresponding frame, acquiring its object-related visual contents.
\newline

\noindent\textbf{Detection Head}.
A feed-forward network with a detection head is append after each decoder layer to iteratively refine the detection results.
Taking an object query $\bm{q}_{ij}$ as input, the detection head adopts a localization branch to predict a box offsets $\delta_{ij}$ to update the reference box ${b}_{ij}$; and a classification branch to predict the class logits ${p}_{ij}$.
Denote by ${y}_{i} = \{(p_{ij}, b_{ij})\}_{j=1}^L$ the predicted objects on the $i$-th frame and $y^*_{i} = \{(c_{ij}, b^*_{ij})\}_{j=1}^L$ the corresponding ground-truth objects padded with $\varnothing$.
Following~\cite{carion2020end}, ClipVID applies the set prediction loss which first finds an optimal bipartite matching between ${y}_{i}$ and ${y}^*_{i}$ by searching for a permutation of $L$ elements $\sigma \in S_L$ with the lowest cost:
\begin{equation}
\sigma = \arg\min_{\sigma \in S_L}\sum_{j=1}^{L}\mathcal{L}_{match}(y^*_{ij}, y_{i\sigma(j)}),
\end{equation}
where $\mathcal{L}_{match}$ is defined as
\begin{equation}
\begin{split}
\mathcal{L}_{match}(y^*_{ij}, {y}_{i\sigma(j)}) & = \lambda_{cls}\mathcal{L}_{cls}(c_{ij}, {p}_{i\sigma(j)}) \\
& + \lambda_{giou}\mathcal{L}_{giou}(b^*_{ij}, {b}_{i\sigma(j)}) \\
    & + \lambda_{L1}\mathcal{L}_{L1}(b^*_{ij}, {b}_{i\sigma(j)}).
\end{split}\label{eq:match}
\end{equation}
Here, $\mathcal{L}_{cls}$ indicates the focal loss \cite{lin2017focal}, $\mathcal{L}_{giou}$ and $\mathcal{L}_{L1}$ are the GIoU loss \cite{rezatofighi2019generalized} and L1 loss, respectively.
$\lambda_{*}$ are coefficients of the loss terms. 
Then, the training objective is defined to have the same form as Eq.~(\ref{eq:match}), but it is only applied to the matched pairs. 
The final loss is the sum of all matched pairs normalized by the number of objects inside the whole video clip.

\subsection{Identity-Consistent Aggregation}
The ICA module is applied in the last several layers of the transformer decoder.
It consists of a Matching step, which introduces an additional identity embedding branch to the detection head of the previous decoder layer;
and an Aggregation step, where an identity-consistent aggregation layer is inserted between the self-attention and cross-attention operations of the current decoder layer.
\newline

\noindent\textbf{Matching}.
The identity embedding branch is a two-layer MLP followed by an $L_2$ normalization layer that projects the object query $\bm{q}_{ij}$ into an object identity embedding $\bm{h}_{ij} \in \mathbb{R}^d$.
Then, given the $n$-th object query on the $m$-th frame, $\bm{q}_{mn}$, we select its most similar object query in each of the rest frames according to their dot-product similarity.
The selected object queries are considered to have the same identity with $\bm{q}_{mn}$, \emph{i.e.}, identity-consistent object queries of $\bm{q}_{mn}$, denoted by $\{\bm{q}_{iJ_{mn}(i)}|i\in I_{m}\}$ where:
\begin{equation}
\begin{split}
J_{mn}(i) &= \arg\max_{j \in [1, L]} \bm{h}_{mn} \cdot \bm{h}_{ij}, \\ \forall i \in I_{m},&~~I_{m} = \{i | i \in [1, T], i \neq m\}.
\end{split}\label{eq:idassign}
\end{equation}

To train the identity embedding branch, 
suppose the set $I^*$ and $J^*$ contain the frame indexes and the query indexes of all object queries that are assigned to the same ground-truth video object according to Eq.~(\ref{eq:match}), respectively. 
Then, any two indexes $i$ and $m$ in $I^*$ will give us a pair of object queries $\bm{q}_{iJ^*(i)}$ and $\bm{q}_{mJ^*(m)}$ that are consistent in their object identity, which should have a relatively small distance in the embedding space.
To achieve this, we use an additional contrastive loss to train the parameters for identity-consistent feature aggregation:
\begin{equation}
\begin{split}
\mathcal{L}_{con} = -&\sum_{\forall m,i \in Z}\log\frac{\exp( \bm{h}_{mJ^*(m)} \cdot \bm{h}_{iJ^*(i)} )}{\sum_{j=1}^L \exp( \bm{h}_{mJ^*(m)} \cdot \bm{h}_{ij} )},\\ &Z = \{m,i|m,i\in I^*, m \neq i\}.
\end{split}\label{eq:contrastive}
\end{equation}
The final contrastive loss is the sum of $\mathcal{L}_{con}$ for all video objects, normalized by the number of all matched pairs.
\newline

\begin{figure}[t]
\centering
\includegraphics[width=0.85\linewidth]{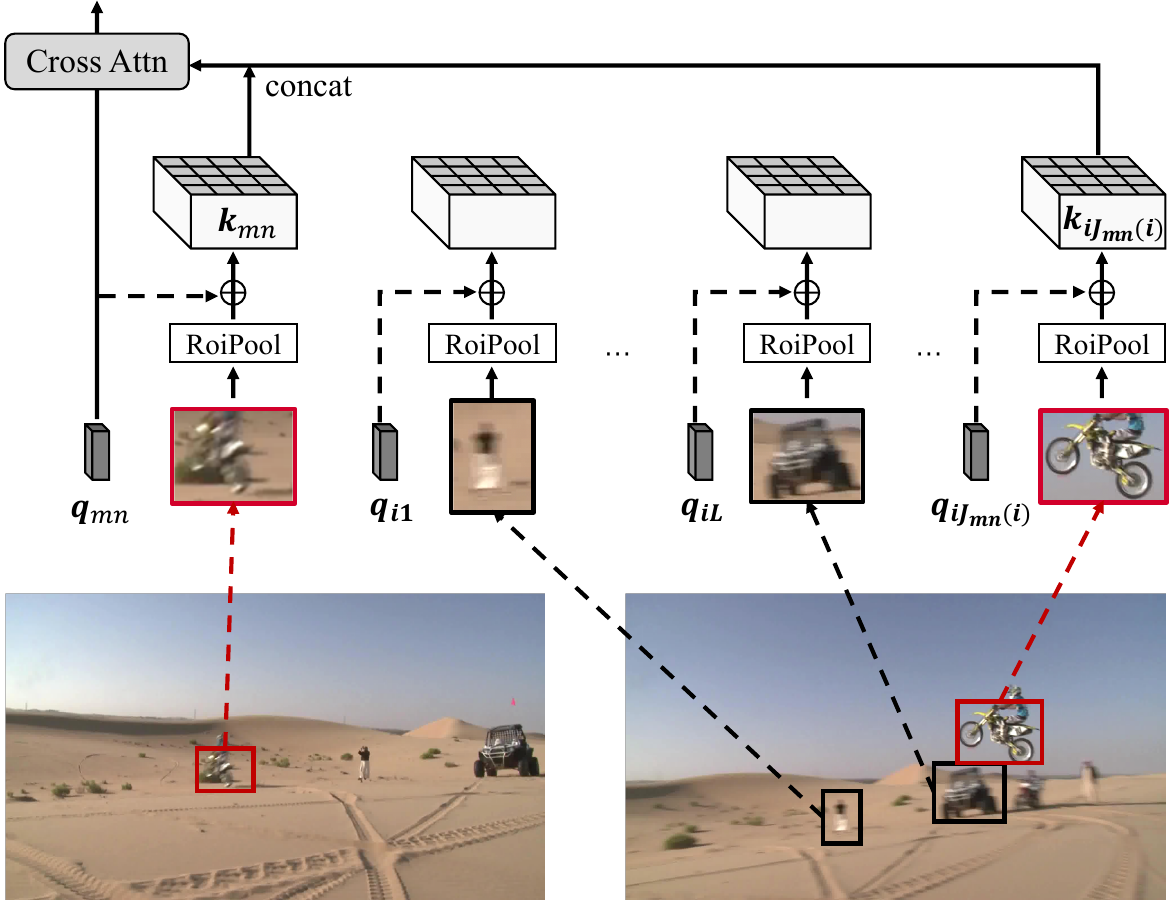}
\caption{
Illustration of the ICA process.
In this example, $\bm{q}_{mn}$ and $\bm{q}_{iJ_{mn}(i)}$ indicate the same motorbike on the $m$-th frame and $i$-th frame, respectively.
Thus, to obtain the ``global view'' of the object query $\bm{q}_{mn}$, only its identity-consistent temporal contexts, \emph{i.e.}, the temporal contexts extracted from $\bm{q}_{iJ_{mn}(i)}$, are used for aggregation.
The dashed lines indicate the positional embedding generated from the object queries.
}
\label{fig:ca}
\end{figure}

\noindent\textbf{Aggregation}.
For $\bm{q}_{mn}$ and its identity-consistent object queries $\{\bm{q}_{iJ_{mn}(i)}|i\in I_{m}\}$, we reuse their corresponding region features $\bm{k}_{mn}$ and $\{\bm{k}_{iJ_{mn}(i)}|i\in I_{m}\}$ obtained in Eq.~(\ref{eqn:roifeat}), and stack them into a joint representation $\bm{K}_{mn} \in \mathbb{R}^{Ts^2\times d}$.
Note that, $\bm{K}_{mn}$ consists of the fine-grained grid-level feature representations of the same video object (ideally) from multiple frames, which is adopted as the identity-consistent temporal contexts for $\bm{q}_{mn}$.
Finally, a cross-attention operation is performed using $\bm{q}_{mn}$ as the query and $\bm{K}_{mn}$ as the source, where $\bm{q}_{mn}$ attends to all $T\times s^2$ elements in $\bm{K}_{mn}$ to obtain its identity-related information at a fine-grained level, results in a more comprehensive ``global view'' of the corresponding video object.
Taking this global view as input, the guided cross-attention layer then retrieves its local view from the region feature $\bm{k}_{mn}$.
A detailed illustration of the ICA process is given in Figure \ref{fig:ca}.

\section{Experiments}
\noindent\textbf{Dataset}. We conduct experiments on the widely used benchmark dataset ImageNet VID \cite{russakovsky2015imagenet}.
It contains 30 object categories and has 3,862 training videos and 555 validation videos.
Mean Average Precision (mAP) is adopted as the evaluation metric.
\newline

\setlength{\tabcolsep}{5pt}
\begin{table}[t]
\centering
\small
\begin{tabular}{@{}l|c|c|c@{}}
\toprule
Methods & Backbone & mAP & fps \\
\midrule
FGFA \cite{zhu2017flow} & R-101 & 76.3 & - \\
PSLA \cite{guo2019progressive} & R-101 & 77.1 & 30.8 \\
MANet \cite{wang2018fully} & R-101 & 78.1 & - \\
STSN \cite{bertasius2018object} & R-101 & 78.9 & - \\
OGEMN \cite{deng2019object} & R-101 & 79.3 & - \\
LRTRN \cite{shvets2019leveraging} & R-101 & 81.0 & 9.5 \\
RDN \cite{deng2019relation} & R-101 & 81.8 & 10.6 \\
TransVOD \cite{he2021end} & R-101 & 81.9 & - \\
SELSA \cite{wu2019sequence} & R-101 & 82.7 & 9.6 \\
MEGA \cite{chen2020memory} & R-101 & 82.9 & 5.8 \\
HVRNet \cite{han2020mining} & R-101 & 83.2 & - \\
TF-Blender \cite{cui2021tf} & R-101 & 83.8 & $< 5.8$ \\
DSFNet \cite{lin2020dual} & R-101 & 84.1 & $< 5.8$ \\
\midrule
*STMM \cite{xiao2018video} + SeqNMS \cite{han2016seq} & R-101 & 80.5 & $\sim 1$ \\
*RDN \cite{deng2019relation} + BLR \cite{deng2019relation} & R-101 & 83.8 & $\sim 1$ \\
*HVRNet \cite{han2020mining} + SeqNMS \cite{han2016seq} & R-101 & 83.8 & $\sim 1$ \\
*MEGA \cite{chen2020memory} + BLR \cite{deng2019relation} & R-101 & 84.5 & $\sim 1$ \\
\midrule
ClipVID & R-101 & \textbf{84.7} & 39.3 \\
\midrule
RDN \cite{deng2019relation} & X-101 & 83.2 & - \\
MEGA \cite{chen2020memory} & X-101 & 84.1 & 5.3 \\
SELSA \cite{wu2019sequence} & X-101 & 84.3 & - \\
HVRNet \cite{han2020mining} & X-101 & 84.8 & - \\
\midrule
*RDN \cite{deng2019relation} + BLR \cite{deng2019relation} & X-101 & 84.7 & $\sim 1$ \\
*MEGA \cite{chen2020memory} + BLR \cite{deng2019relation} & X-101 & 85.4 & $\sim 1$ \\
*HVRNet \cite{chen2020memory} + SeqNMS \cite{han2016seq} & X-101 & 85.5 & $\sim 1$ \\
\midrule
ClipVID & X-101 & \textbf{85.8} & {25.1} \\
\bottomrule
\end{tabular}
\caption{Comparisons with state-of-the-art methods on ImageNet VID dataset. * indicates the use of sequence-level post-processing methods like SeqNMS and BLR. 
}\label{tab:sota}
\end{table}

\noindent\textbf{Implementation Details}.
We use ResNet-101 \cite{he2016deep} with dilated convolutions \cite{chen2017deeplab} in the last stage as the backbone for analysis.
The transformer decoder has 6 layers, 8 attention heads, and a hidden dimension of $d=384$.
The number of object queries for each frame is set to 72. 
The reference boxes are initialized as the frame size, and the output size of the RoIPooling operation is 7.
By default, the identity-consistent feature aggregation is only performed in the last two decoder layers, within the top-$10$ object queries that have the largest classification scores of each frame.
$\lambda_{cls}$, $\lambda_{giou}$, and $\lambda_{L1}$ are set to 2, 2, 5, respectively, as in~\cite{carion2020end}. 

Following the common practice \cite{chen2020memory,deng2019relation}, we utilize both ImageNet VID and ImageNet DET datasets to train our model. 
During training, we randomly sample $T=3$ frames from the same video.
During inference, we use $T=30$ frames by default.
The frames are resized to a shorter side of 600 pixels.
The backbone network is initialized with the ImageNet pre-trained weights, the rest model parameters are randomly initialized.
The training process is separated into two stages.
In the first stage, we train all model parameters except those used by the ICA modules for 180k iterations using the AdamW \cite{loshchilov2018decoupled} optimizer with a total batch size of~4.
The initial learning rate is set to 1e-5 and is divided by 10 at the 120k-th iteration.
We then train the whole model for another 60k iterations, using an initial learning rate of 1e-6, and divide it by 10 at the 40k-th iteration.

\setlength{\tabcolsep}{5pt}
\begin{table}[t]
\centering
\small
\resizebox{1\linewidth}{!}{
\begin{tabular}{@{}l|c|c|c|c@{}}
\toprule
Methods & mAP(\%) & slow (\%) & medium (\%) & fast (\%)\\
\midrule
SELSA \cite{wu2019sequence} & 82.7 & 88.0 & 81.4 & 67.1 \\
MEGA \cite{chen2020memory} & 82.9 & 89.4 & 81.6 & 62.7\\
HVRNet \cite{han2020mining} & 83.2 & 88.7 & 82.3 & 66.6\\
DSFNet \cite{lin2020dual} & 84.1 & 90.0 & 82.6 & 67.0  \\
\midrule
ClipVID w/o ICA & 83.3 & 89.0 & 82.7 & 66.1 \\
ClipVID  & 84.7 & 89.9 & 83.9 & 68.5 \\
ClipVID (oracle ICA)  & 85.8 & 90.8 & 84.5 & 72.3 \\
\bottomrule
\end{tabular}
}
\caption{Detailed performance comparisons on ImageNet VID. "slow/medium/fast" indicates the mean Average Precision for video objects with slow/medium/fast moving speed.}\label{tab:adq}
\end{table}

\setlength{\tabcolsep}{3pt}
\begin{table}[t]
\centering
\small
\resizebox{1\linewidth}{!}{
\begin{tabular}{@{}c|c|c|c|cc|cc|cc@{}}
\toprule
Encoder & Adaptive  & Extend  & Guided  & \multicolumn{2}{c|}{1 frame} & \multicolumn{2}{c|}{5 frames} & \multicolumn{2}{c}{30 frames}\\ \cmidrule(l){5-10} 
Free& Query  &    SA       &     CA       &  mAP &  fps &  mAP &  fps &  mAP & fps  \\ \midrule
&   &           &            & 61.8 & 24.4 & \multicolumn{2}{c|}{N/A} & \multicolumn{2}{c}{N/A} \\ 
\checkmark &    &            & \checkmark & 76.7 & 45.2 & \multicolumn{2}{c|}{N/A} & \multicolumn{2}{c}{N/A} \\ 
&   &\checkmark &            & 63.2 & 24.4 & 63.5 & 14.2 & \multicolumn{2}{c}{OOM} \\ 
\checkmark &   &\checkmark & \checkmark & 78.1 & 43.5 & 79.6 & 43.1 & 82.4 & 41.9 \\
\checkmark & \checkmark &\checkmark & \checkmark & 78.3 & 43.5 & 80.1 & 43.1 & 83.3 & 41.7 \\
\bottomrule
\end{tabular}
}
\caption{Performance analysis based on ClipVID w/o ICA. ``SA'' and ``CA'' are short for Self-Attention and Cross-Attention, respectively. The first row and the last row indicate the performance of the original DETR and the ClipVID w/o ICA, respectively.
The mAP values are shown in percentage.}\label{tab:ClipVID}
\end{table}

\subsection{Comparisons with state of the arts}
We first compare the proposed ClipVID with previous SOTA methods.
As shown in Table \ref{tab:sota}, ClipVID achieves significantly faster (about $7\times$) inference speed than recent SOTA methods like TF-Blender \cite{cui2021tf} and DSFNet \cite{lin2020dual}, while also outperforms them by a large margin in terms of mAP, \emph{e.g.}, 84.7\% vs. 84.1\%.
Compared with PSLA \cite{guo2019progressive}, a VID model specifically optimized for efficient inference in real-world scenarios, our model still outperforms its speed clearly (39.3 fps vs 30.8 fps).
More importantly, PSLA uses sparse computation techniques to reduce computation, which is also applicable to our model to further improve the inference speed.
In terms of performance, ClipVID outperforms PSLA by 6.2\%, making it a better choice for deployment.
Note that, the proposed ClipVID model is fully end-to-end trainable.
Still, it outperforms previous SOTA approaches that are equipped with sequence-level post-processing techniques like Seq-nms~\cite{han2016seq} and BLR \cite{deng2019relation}, while being nearly 40 times faster\footnote{measured based on our implementations.}.
Compared with TransVOD~\cite{he2021end}, an end-to-end VID model, our model obtains a much better performance (84.7\% vs. 81.9\%) with a much simpler network architecture.
We obtain similar observations when using a stronger backbone ResNext-101-32x8d for ClipVID, where it outperforms SOTA methods in terms of both speed and accuracy.
These results demonstrate the effectiveness of the proposed method.

\subsection{Performance Analysis of ClipVID}
We first analyze the performance of the proposed Identity-consistent Aggregation approach.
As shown in Table \ref{tab:adq}, there is a clear performance degradation (from 84.7\% to 83.3\%) when removing ICA from the ClipVID model.
This degradation is more obvious for fast-moving objects, where mAP fast drops significantly by 2.4\%.
Moreover, compared with the previous method, our ClipVID is generally much more accurate in detecting fast-moving objects.
These results show the effectiveness of our ICA module in handling large appearance variations.
Also note that ClipVID w/o ICA is only slightly faster than ClipVID, indicating that applying identity-consistent aggregation in ClipVID is very efficient.
We further conduct an experiment using the oracle matching process, \ie, for each object query, we find its identity-consistent object query by choosing those that are assigned to the same video object.
We find that the performance improves clearly with the oracle query matching, especially for the fast-moving objects, where the mAP is booted by $3.8\%$.

\setlength{\tabcolsep}{4pt}
\begin{table}[t]
\centering
\resizebox{1\linewidth}{!}{
\begin{tabular}{@{}l|cc|cc|cc}
\toprule
\multirow{2}{*}{Methods} & \multicolumn{2}{c|}{ClipVID} & \multicolumn{2}{c|}{Sparse RCNN} & \multicolumn{2}{c}{Deformable DETR} \\
     &  w/ ICA  &  w/o ICA   & w/ ICA  &  w/o ICA  &  w/ ICA  & w/o ICA \\ \midrule
mAP (\%) &  84.7  &  83.3  &  84.1  &  82.5  &  83.9  &  82.2 \\
FPS  & 39.3   &  41.7  &  35.0  &  36.8  &  33.1  &  34.7 \\ \bottomrule
\end{tabular}
}
\caption{Results of ClipVID with different query-based object detectors.}\label{tab:detector}
\end{table}

\setlength{\tabcolsep}{4pt}
\begin{table}[t]
\centering
\small
\begin{tabular}{@{}l|cccc@{}}
\toprule
Method  & Last layer & Last 2 layers & Last 3 layers & All\\ \midrule
mAP (\%)    & 84.4 & 84.7 & 84.5 & 83.9\\ \bottomrule
\end{tabular}
\caption{Performance analysis on the number of decoder layers that adopt the Identity-consistent Aggregation module.}\label{tab:ca1}
\end{table}

\begin{table}[t]
\small
\centering
\begin{tabular}{@{}l|cccc}
\toprule
Method  & Top-5 & Top-10 & Top-15 & All  \\ \midrule
mAP (\%)    & 84.2 & 84.7 & 84.5 & 83.6 \\
fps    & 39.8 & 39.3 & 38.5 & 30.3  \\ \bottomrule
\end{tabular}
\caption{Performance and speed analysis on the number of object queries that perform the identity-consistent temporal context aggregation. Top-$k$ denotes selecting the $k$ object queries with the highest classification scores.
}\label{tab:ca2}
\end{table}

Besides the proposed ICA module, ClipVID makes several modifications to the DETR framework, including 1) removing the transformer encoder; 2) using adaptive object queries; 3) extending the self-attention across frames; and 4) using guided cross-attention.
Here, we analyze how our model benefits from them in Table \ref{tab:ClipVID}.
Note that we use ClipVID w/o ICA as the base model to better reveal the contributions of these modifications.
From the table, the first two rows indicate performing image-level object detection without considering the temporal contexts. 
We find that the original DETR (the first row) produces much worse performance compared to its ``encoder-free + guided cross-attention'' counterpart (row 2), which may be largely due to its low convergence speed.
Besides, the encoder also slows down the inference speed by nearly 2 times.
In terms of video object detection, DETR trained with extended self-attention (row 3) improves the performance clearly over its original results in the single-frame inference setting.
In the multi-frame setting, however, the performance is only improved marginally (from 63.2\% to 63.5\%) when using 5 inference frames, while suffering from a severe slowdown of the inference speed.
We believe that the extremely-long feature sequences in the video setting not only dramatically increase the computation complexity of the encoder but also hampers the learning of the encoder and prevent it from leveraging useful information from the temporal contexts. 
Moreover, using more inference frames failed in this setting due to memory limitation.

\begin{table}[t]
\small
\centering
\begin{tabular}{@{}l|cccccc@{}}
\toprule
\# infer frames  & 1 & 10 & 15 & 25 & 30 & 45 \\ \midrule
mAP (\%)      & 78.3 & 82.1 & 82.7  & 83.2 & 83.3 & 83.4 \\ 
fps     &  43.5 & 42.7 &  42.4  &  41.9 & 41.7 & 37.3  \\ 
\bottomrule
\end{tabular}
\caption{Performance and speed analysis on the number of inference frames in ClipVID w/o ICA.}\label{tab:frames}
\end{table}

\setlength{\tabcolsep}{7pt}
\begin{table}[t]
\centering
\small
\begin{tabular}{@{}l|ccccc@{}}
\toprule
\# object queries & 48 & 64 & 72 & 80 & 100 \\ \midrule
mAP (\%)        & 80.6 & 82.4 & 83.3 & 82.9 & 82.1 \\
fps        & 45.7 & 44.4 & 41.7 & 40.0 & 37.2  \\ \bottomrule
\end{tabular}
\caption{Performance and speed analysis on the number of object queries per frame in ClipVID w/o ICA.}\label{tab:queries}
\end{table}

\setlength{\tabcolsep}{10pt}
\begin{table}[t]
\small
\centering
\begin{tabular}{@{}l|cccc@{}}
\toprule
\# decoder layers & 4 & 5 & 6 & 7\\ \midrule
mAP (\%)   & 81.9 & 82.8 & 83.3 & 83.3\\
fps        & 45.6 & 44.1 & 41.7 & 38.0 \\ \bottomrule
\end{tabular}
\caption{Performance and speed analysis on the number of decoder layers in ClipVID w/o ICA.}\label{tab:layers}
\end{table}

\begin{figure*}[t]
\centering
\includegraphics[width=0.95\linewidth,height=0.44\textheight]{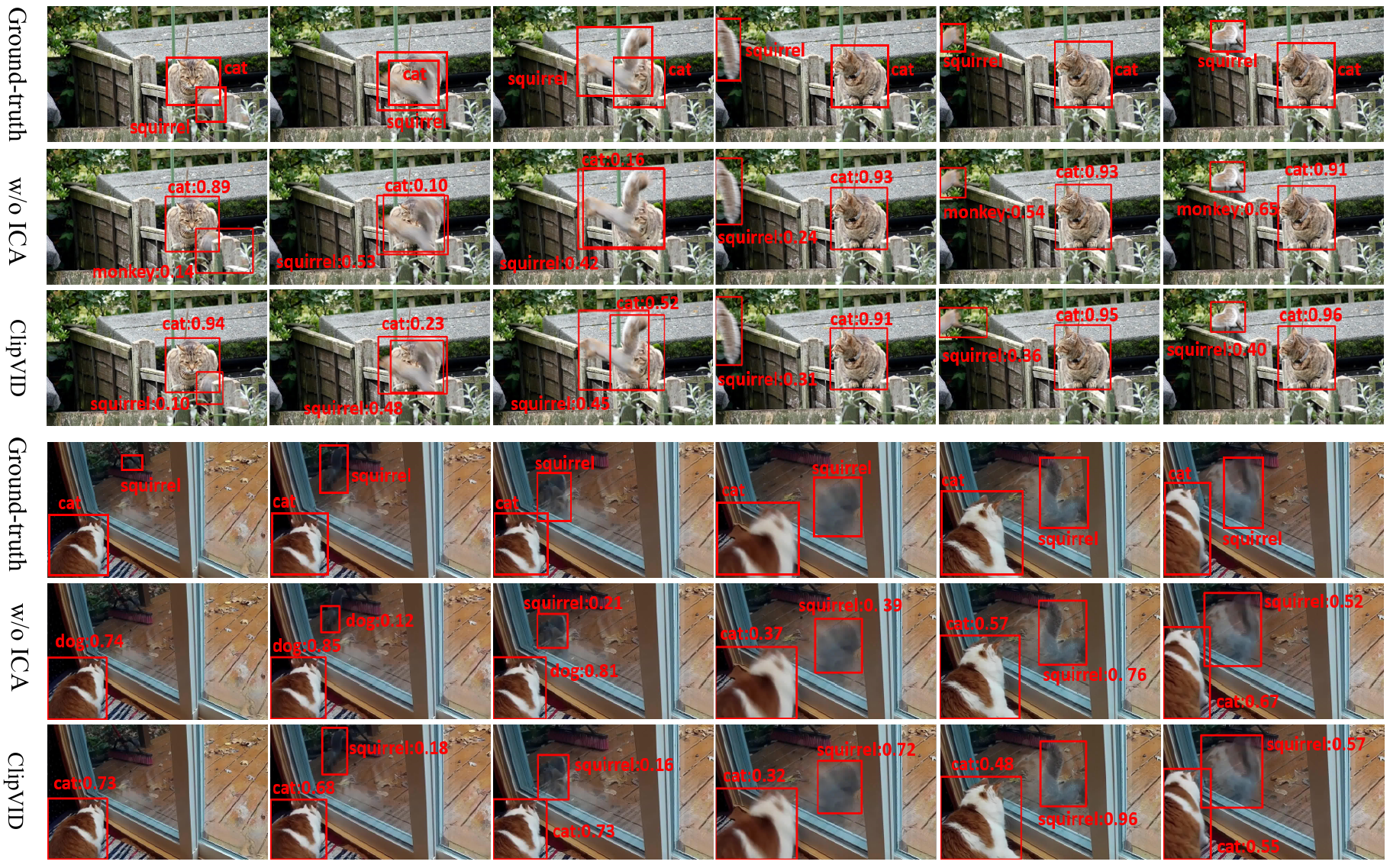}
\caption{Visualization of the detection results. ``w/o ICA'' indicates the ClipVID w/o ICA model. For simplicity, we only show the detection results with confidence scores higher than 0.1.}
\label{fig:vis}
\end{figure*}

On the other hand, our ClipVID w/o ICA model (the last row) enjoys having more inference frames, where its performance is boosted from 78.3\% in the single-frame setting to 80.1\% in the 5-frame setting and is further increased significantly to 83.3\% when using 30 inference frames.
This observation verifies the effectiveness of extended self-attention in propagating temporal information.
Meanwhile, the inference speed of our model is consistently fast (more than 40 fps) in all inference settings thanks to the elimination of the transformer encoder and its clip-wise prediction manner.
Comparing the last two rows shows that the adaptive object queries perform better than fixed object queries, especially when the number of input frames is large.

Lastly, we show the performance of using different query-based object detection methods for ClipVID in Table~\ref{tab:detector}.
We consider Sparse RCNN~\cite{sun2021sparse} and Deformable DETR~\cite{zhu2020deformable}, which are widely used and have shown strong performance in still image object detection.
However, we find that they perform inferior to our simple detector modified from DETR, and are also slightly slower.
Notably, both Sparse RCNN and Deformable DETR adopt a heavy decoder (\ie, having more parameters and incurring more computation cost) which may lead to overfitting in the relatively small ImageNet VID dataset.


\subsection{Hyper-parameter Analysis}
In this section, we analyze the hyper-parameters in our model design.
We first analyze the effect of applying the Identity-consistent Aggregation for different numbers of decoder layers (Table \ref{tab:ca1}), as well as the effect of using it for different numbers of object queries (Table \ref{tab:ca2}).
From Table \ref{tab:ca1}, our model achieves the best result when applying ICA in the last two decoder layers.
Moreover, using it for the last one or three decoder layers yields similar performances to the best setting.
When using the ICA module for all decoder layers, the performance degrades to 83.9, showing that the early decoder stages may not be able to capture the object identity.
From Table \ref{tab:ca2}, applying ICA on the top-5, top-10, and top-15 object queries all bring clear gains over the baseline model ClipVID w/o ICA.
Among them, choosing the top-10 scored object queries to perform ICA achieves the best result.
When applying identity-consistent temporal context aggregation for all object queries, a clear performance degradation is observed.
We hypothesize that the object queries with low classification scores may hamper the matching step of ICA and further introduce noises to the temporal contexts.
Besides, the inference speed is also decreased clearly in this setting.

We provide more hyper-parameter analysis based on the ClipVID w/o ICA model.
Specifically, we analyze how the number of inference frames per input clip influences the performance.
From Table \ref{tab:frames}, our model benefits from larger temporal receptive fields, which is aligned with previous works.
Thanks to the clip-wise prediction manner of our method, the inference speed is only mildly reduced when increasing the number of inference frames from 1 to 45.
Then, we study the effect of the number of object queries per frame on the model performance.
As shown in Table \ref{tab:queries}, with 72 object queries, our model yields the best performance.
This is different from DETR which benefits from having more than 100 object queries.
The reason could be that the MSCOCO \cite{lin2014microsoft} dataset used by DETR is much more complex than ImageNet VID, in terms of the object categories and the number of objects per image.
Thus, DETR requires more object queries to increase its object representation capacity.
Lastly, we show the performance of our model with different numbers of transformer decoder layers in Table \ref{tab:layers}.
From the table, 6 decoder layers are sufficient for ClipVID to achieve strong performance.
Using fewer decoder layers can increase the inference speed but at the cost of clear performance degradation. 

\subsection{Visualization}
\label{sec_vis}
We further visualize some detection results of the proposed methods in Figure \ref{fig:vis}.
From the figures, the proposed ICA module qualitatively improves the detection performance.
On some hard cases, ClipVID w/o ICA fails to make accurate predictions, \emph{e.g.},
in the first column and last two columns of the first example, ClipVID w/o ICA mistakenly recognizes the squirrel as a monkey due to occlusion and the unusual pose of the squirrel.
Besides, for some objects, ClipVID w/o ICA makes low confident predictions, like the cat in the third column of the first example and the squirrel in the fourth column of the second example, due to occlusion and motion blur, respectively.

\section{Conclusion}
Existing VID models usually treat the temporal contexts from different video objects indiscriminately despite their different identities.
This may hamper the learning of object representations due to the irrelevant and noisy information contained in the temporal contexts.
In this paper, we aim to perform Identity-Consistent temporal context Aggregation (ICA) to enhance the video object representations.
To achieve this, we first need to reduce the redundancies in the temporal context so that ICA can be done efficiently.
Thus, we proposed a VID model called ClipVID which is based on the DETR framework.
ClipVID is able to perform identity-consistent aggregation, while also 
effectively removing the redundancies and 
making predictions for all input frames simultaneously, making the model very efficient.
In the experiment, our ClipVID model outperforms previous SOTAs on the benchmark ImageNet VID dataset in terms of both speed and accuracy.

{\small
\bibliographystyle{ieee_fullname}
\bibliography{egbib}
}

\end{document}